\documentclass[10pt,twocolumn,letterpaper]{article}

\usepackage{cvpr}
\usepackage{times}
\usepackage{epsfig}
\usepackage{graphicx}
\usepackage{amsmath}
\usepackage{amssymb}
\usepackage[table]{xcolor}

\DeclareMathOperator*{\argmin}{arg\,min}
\usepackage{booktabs}

\usepackage[breaklinks=true,bookmarks=false]{hyperref}

\cvprfinalcopy 

\def\cvprPaperID{7329} 

\newcommand{\afterfigure}{\vspace{-1em}}

\begin{document}

\newcommand{\tali}[1]{{\textcolor{violet}{[Tali: #1]}}}
\newcommand{\ariel}[1]{{\textcolor{green}{[Ariel: #1]}}}
\newcommand{\miki}[1]{{\textcolor{blue}{[Miki: #1]}}}
\newcommand{\oran}[1]{{\textcolor{orange}{[Oran: #1]}}}
\newcommand{\todo}[1]{{\textcolor{red}{[TODO: #1]}}}
\newcommand{\sagie}[1]{{\color{orange}{[Sagie: #1]}}}

\title{\vspace{-0.6in} SpeedNet: Learning the Speediness in Videos}
\author{Sagie Benaim$^{1,2}$\thanks{Performed this work while an intern at Google.} \quad Ariel Ephrat$^1$ \quad Oran Lang$^1$ \quad Inbar Mosseri$^1$ \quad William T. Freeman$^1$\\ Michael Rubinstein$^1$ \quad Michal Irani$^{1,3}$ \quad Tali Dekel$^1$ \\ \\
$^1$Google Research\quad $^2$Tel Aviv University\quad $^3$Weizmann Institute\\}


\maketitle
\begin{abstract}
We wish to automatically predict the ``speediness'' of moving objects in videos---whether they move faster, at, or slower than their ``natural'' speed. 
The core component in our approach is SpeedNet---a novel deep network trained to detect if a video is playing at normal rate, or if it is sped up. SpeedNet is trained on a large corpus of natural videos in a self-supervised manner, without requiring any manual annotations. We show how this single, binary classification network can be used to detect arbitrary rates of speediness of objects.
We demonstrate prediction results by SpeedNet on a wide range of videos containing complex natural motions, and examine the visual cues it utilizes for making those predictions. Importantly, we show that through predicting the speed of videos, the model learns a powerful and meaningful space-time representation that goes beyond simple motion cues. We demonstrate how those learned features can boost the performance of self-supervised action recognition, and can be used for video retrieval. Furthermore, we also apply SpeedNet for generating time-varying, adaptive video speedups, which can allow viewers to watch videos faster, but with less of the jittery, unnatural motions typical to videos that are sped up uniformly.
\end{abstract}

\begin{figure}[t]
    \centering
    \includegraphics[width=\columnwidth]{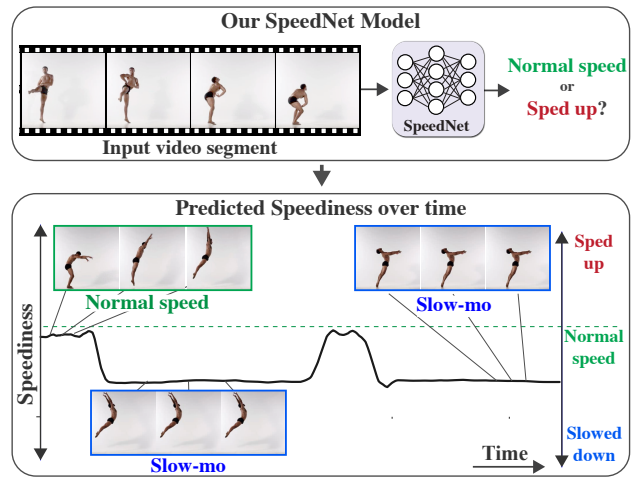}
    \caption{Given an input video, our method automatically predicts the ``speediness'' of objects in the video---whether they move faster, at, or slower than their natural speed. Bottom: a video of a dancer alternates between normal speed and slow motion playback, as correctly captured by our speediness prediction over time. Note that {\bf speediness $\neq$ magnitude of motions}, see Fig.~\ref{fig:speediness_neq_motion_magnitude}. The core component of our method is SpeedNet (top)---a novel deep network that detects whether an object is moving at its normal speed, or faster than its normal speed.}
    \label{fig:teaser}
\afterfigure    
\end{figure}


\vspace{-1em}
\section{Introduction}
\label{sec:intro}

A human observer can often easily notice if an object's motion is sped up or slowed down. For example, if we play a video of a dancer at twice the speed ($2\times$), we can notice unnatural, fast and jittery motions. In many cases, we have prior knowledge about the way objects move in the world (people, animals, cars); we know their typical dynamics and their natural \emph{rate of motion}. 

In this paper, we seek to study how well we can train a machine to learn such concepts and priors about objects' motions. Solving this task requires high-level reasoning and understanding of the way different objects move in the world. We achieve this by training a single model, \emph{SpeedNet}, to perform a basic binary classification task: estimate whether an object in an input video sequence is moving at its normal speed, or faster than the normal speed (Fig.~\ref{fig:teaser}, top). That is, given a set of $L$ frames in an $L$-fps video as input, we set to predict whether those frames depict $1$ second of the object's motion in the world (normal speed), or more than $1$ second (the object/video is sped up). We preferred this approach over directly predicting (regressing to) the playback rate of the video, because our ultimate goal is to determine whether motion in a given video is natural or not, a task for which a regression objective may be unnecessarily difficult to learn.

We dub this the ``speediness'' of the object. We then show how this basic binary classification model can be applied at test time to predict arbitrary rates of speediness in videos, when objects are sped-up, or slowed-down, by different factors (Fig.~\ref{fig:teaser}, bottom). The model is trained on Kinetics \cite{kay2017kinetics}, a large corpus of natural videos of human actions, in a self-supervised manner, without requiring manual labels.

\begin{figure}[t]
    \hspace*{-.23in}
    \includegraphics[width=1.05\columnwidth]{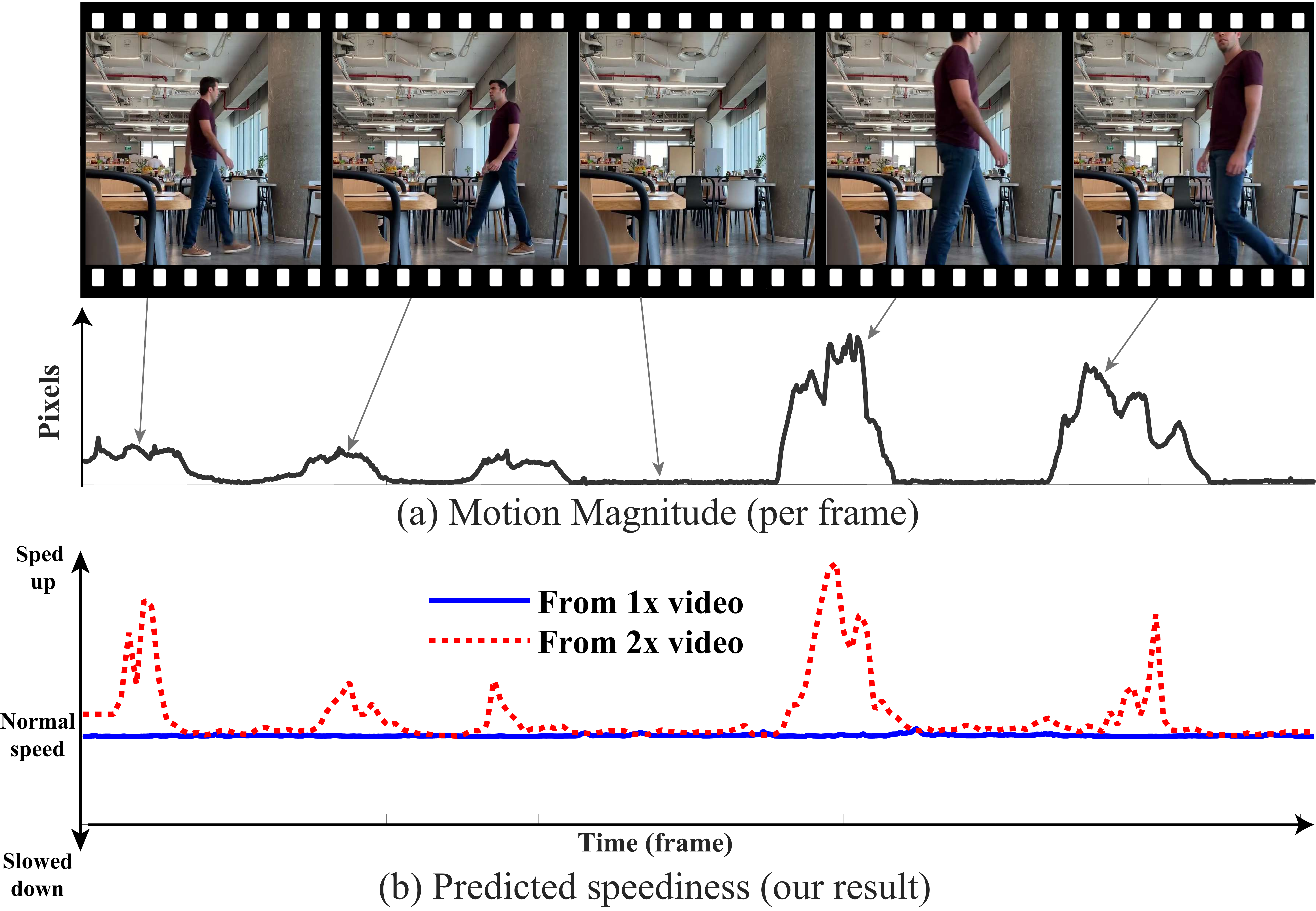}
    \caption{{\bf Speediness $\neq$ motion magnitude.} A person is walking back and forth, at first further away from the camera, then closer to the camera (top). The magnitude of motions varies significantly throughout the sequence (in particular, the motions get larger when the person is closer to the camera; middle plot), but our SpeedNet model is able to produce a stable classification of normal speed throughout the video (speediness score close to zero). If we input to SpeedNet the video played at twice the speed ($2 \times$), then the walking segments do indeed get recognized as faster-than-normal human motions (higher speediness scores), whereas the static segments (no person in the scene), are classified as normal speed.}
    \label{fig:speediness_neq_motion_magnitude}
\afterfigure    
\end{figure}

Our motivation for this study is twofold. First, we ask if it is possible to train a reliable classifier on a large-scale collection of videos to predict if an object's motion is sped up or played at normal speed---what would such a model learn about the visual world in order to solve this task? Second, we show that a well-trained model for predicting speediness can support a variety of useful applications.

Training SpeedNet, however, is far from trivial. In humans, the ability to correctly classify an object's speed continues to improve even throughout adolescence \cite{manning2012development}, implying that a developed mind is required to perform this task. In addition to the high-level reasoning necessary for solving this task, a major challenge in training a neural network to automatically predict speediness is to avoid their tendency to detect easy shortcuts, e.g., to rely on artificial, low-level cues, such as compression artifacts, whenever possible. This usually results in near-perfect accuracy on the learning task, which is an outcome we want to avoid: we seek a semantic, as opposed to artificial, understanding of speediness, by examining the \emph{actual} motions, and not relying on artifacts (compression, aliasing) related to the way in which the sped-up videos are generated. We describe the strategies employed to mitigate artificial shortcuts in Sec.~\ref{sec:cues}.

Another challenging aspect of training SpeedNet is to go beyond the trivial case of motion magnitude for determining speediness. Relying only on the speed of moving objects, using motion magnitude alone, 
would discriminate between, e.g. two people walking normally at two different distances from the camera (Fig.~\ref{fig:speediness_neq_motion_magnitude}). We address the speediness prediction capability of optical flow in Sec.~\ref{sub:speedup_predictions}, and demonstrate the clear prediction superiority of SpeedNet over a naive flow-based baseline method. However, the correlation between speediness and motion magnitude does pose a formidable challenge to our method in cases of extreme camera or object motion.

We demonstrate our model's speediness prediction results on a wide range of challenging videos, containing complex natural motions, like dancing and sports. We visualize and examine the visual cues the model uses to make those predictions.
We also apply the model for generating time-varying, adaptive speedup videos, speeding up objects more (or less) if their speediness score is low (or high), so that when sped up, their motion looks more natural to the viewer. This is in contrast to traditional video speedup, e.g. in online video streaming websites, which use uniform/constant speedup, producing unnatural, jittery motions. Given a desired global speedup factor, our method calculates a smooth curve for per-frame speedup factor based on the speediness score of each frame. The details of this algorithm are described in Sec.~\ref{sec:adaptive_speedup}.


Lastly, we also show that through learning to classify the speed of a video, SpeedNet learns a powerful space-time representation that can be used for self-supervised action recognition and for video retrieval. Our action recognition results are competitive with the state-of-the-art self-supervised methods on two popular benchmarks, and beat all other methods which pre-train on Kinetics. We also show promising results on cross-video clip retrieval.

Our videos, results, and supplementary material are available on the project web page: \url{http://speednet-cvpr20.github.io}.

\section{Related work}

\paragraph{Video playback speed classification.}
Playback speed classification is considered a useful task in itself, especially in the context of sports broadcasts, where replays are often played at different speeds. A number of works try to detect replays in sports \cite{wang2004generic, chen2015novel, javed2016efficient, kiani2012effective}. However, these works usually employ a specific domain analysis, and use a supervised approach. Our method, however, works on any type of video and does not use any information unique to a specific sport. To the best of our knowledge, there exists no public dataset for detection of slow motion playback speed.

\vspace{-1em}
\paragraph{Video time remapping.}
Our variable speedup technique produces non-uniform temporal sampling of the video frames. This idea was explored by several papers. The early seminal work of Bennett and McMillan~\cite{bennett2007computational} calculates an optimal non-uniform sampling to satisfy various visual objectives, as captured by error metrics defined between pairs of frames. Zhou \emph{et al.}~\cite{zhou2014time} use a measure of ``frame importance'' based on motion saliency estimation, to select important frames. Petrovic \emph{et al.}~\cite{petrovic2005adaptive} perform query-based adaptive video speedup---frames similar to the query clip are played slower, and different ones faster. An important task is of intelligent fast-forwarding of regular videos \cite{lan2018ffnet}, or egocentric videos \cite{poleg2015egosampling, silva2019semantic, furlan2018fast, silva2018weighted}, where frames are selected to preserve the gist of the video, while allowing the user to view it in less time. All of these works try to select frames based on a saliency measure to keep the maximal ``information'' from the original video, or minimal camera jitter in the case of first-person videos. In contrast, our work focuses on detecting regions which are played at slower than their natural speed. This allows for optimizing a varying playback rate, such that speedup artifacts of moving objects (as detected by the model) will be less noticeable.

\vspace{-1em}
\paragraph{Self-supervised learning from videos.}
Using video as a natural source of supervision has recently attracted much interest \cite{jing2019self}, and many different video properties have been used as supervision, such as: cycle consistency between video frames \cite{wang2019learning, Dwibedi_2019_CVPR};  distinguishing between a video frame sequence and a shuffled version of it \cite{misra2016shuffle, fernando2017self, xu2019self}; solving space-time cubic puzzles \cite{kim2019self}. Another common task is predicting the future, either by predicting pixels of future frames \cite{mathieu2015deep, diba2019dynamonet}, or an embedding of a future video segment \cite{han2019video}. Ng \emph{et al.}~\cite{ng2018actionflownet} try to predict optical flow, and Vondrick~\emph{et al.}~\cite{vondrick2018tracking} use colorization as supervision.

The works most related to ours are those which try to predict the arrow of time \cite{wei2018learning, pickup2014seeing}. This task can be stated as classifying the playback speed of the video between $-1$ and $+1$, as opposed to our work which attempts to discriminate between different positive video speeds.  In concurrent work, Epstein \emph{et al.}~\cite{epstein2019oops} leverage the intrinsic speed of video to predict unintentional action in videos.

\section{SpeedNet} \label{sec:method}

The core component of our method is \emph{SpeedNet}---a deep neural network designed to determine whether  objects  in  a  video  are  moving  at,  or  faster than, their  normal  speed. As proxies for natural and unnaturally fast movement, we train SpeedNet to discriminate between videos played at normal speed and videos played at twice ($2\times$) their original speed. More formally, the learning task is: given a set of $L$ frames extracted from an $L$-fps video as input, 
SpeedNet predicts if those frames contain $1$ second of movement in the world (i.e., \emph{normal speed}), or $2$ seconds (i.e., \emph{sped-up}).

It is important to note that videos played at twice the original speed do not always contain unnatural motion. For example, slow walking sped up to fast walking can still look natural. Similarly, when nothing is moving in the scene, a video played at $2\times$ will still show no motion. Thus, the proxy task of discriminating between $1\times$ and $2\times$ speeds does not always accurately reflect our main objective of predicting speediness. Consequently, we do not expect (or desire for) our model to reach perfect accuracy. Moreover, this property of network ambiguity in cases of slow vs. fast natural motion is precisely what facilitates the downstream use of SpeedNet predictions to ``gracefully'' speed up videos.

We describe and demonstrate how we can use this model for predicting the speediness of objects in natural videos played at arbitrary speeds. The motivation for solving this binary classification problem rather then directly regressing to the video's playback rate is because our goal is to determine whether or not the motion in a given video is natural, a task for which a regression objective may be unnecessarily difficult to learn. Moreover, discriminating between two different speeds is more natural for humans as well. We next describe the different components of our framework.

\subsection{Data, supervision, and avoiding artificial cues}
\label{sec:cues}
SpeedNet is trained in a self-supervised manner, without requiring any manually labeled videos. More specifically, our training and testing sets contain two versions of every video segment, a normal speed version, and a sped-up version constructed by temporally subsampling video frames.

 Previous work has shown networks have a tendency to use shortcuts---artificial cues present in the training data, to help them solve the task at hand \cite{wei2018learning, han2019video, doersch2015unsupervised}. 
 Our network too is susceptible to these cues, and we attempt to avoid potential shortcuts by employing the following strategies:

\vspace{-1em}
\paragraph{Spatial augmentations.} 
Our base network, defined in Sec.~\ref{sec:arch}, is fully convolutional, so its input can be of arbitrary dimensions. During training, we randomly resize the input video clip to a spatial dimension $N$ of between $64$ and $336$ pixels. The blurring which occurs during the resize process can help mitigate potential pixel intensity jitter caused by MPEG or JPEG compression of each frame. After passing the input through the base network, we perform spatial global max pooling over the regions in the resulting space-time features. Since the input is of variable size, these regions correspond to differently sized regions in the original, unresized input. This forces our network not to rely only on size-dependent factors, such as motion magnitude.

\vspace{-1em}
\paragraph{Temporal augmentations.} 
We would like to sample a video at either normal speed or twice its normal speed. To introduce variability in the time domain, for normal speed we sample frames at a rate of $1\times$-$1.2\times$ and for the sped-up version we sample at $1.7\times$-$2.2\times$. In more detail, we choose $3T$ consecutive frames from a given video. For normal speed, we randomly pick a skip factor $f$ between $1-1.2$ and skip frames with probability $1 - 1/f$. We then choose $T$ consecutive frames from the remaining frames. For the sped-up version, $f$ is chosen between $1.7-2.2$.

\vspace{-1em}
\paragraph{Same-batch training.} 
For each clip (of $3T$ consecutive frames), we construct a normal speed and a sped-up video, each of length $T$, in the manner described above. We train our model such that each batch contains both normal-speed and sped-up versions of each video clip. We found that this way, our network is significantly less reliant on artificial cues. We note the same type of training was found to be critical in other self-supervised works such as \cite{gidaris2018unsupervised}. See Tab.~\ref{tb:ablation} and the discussion in Sec.~\ref{sub:speedup_predictions} for the quantitative effect of these augmentation strategies.

\subsection{SpeedNet architecture}
\label{sec:arch}

\begin{figure}
    \centering
    \includegraphics[width=0.47\textwidth]{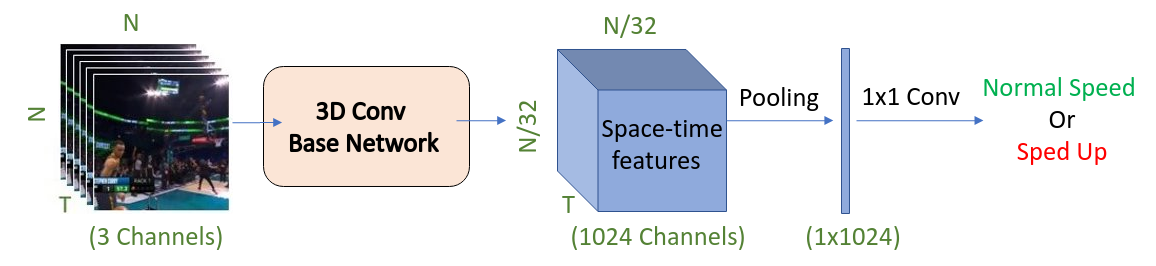}
    \vspace{0.2em}
    \caption{{\bf SpeedNet architecture.} SpeedNet, the core model in our technique, is trained to classify an input video sequence as either normal speed, or sped up. Full details are provided in Sec.~\ref{sec:arch}.} 
    \label{fig:speednet}
\afterfigure
\end{figure}

Our architecture is illustrated in Fig.~\ref{fig:speednet}. The input to the network is a $T \times N \times N$ video segment, which is either sampled from a normal speed video, or its sped-up version ($T$ and $N$ denote the temporal and spatial dimensions, respectively). The input segment is then passed to a fully convolutional base network that learns space-time features. The dimensions of the output features are $T \times \frac{N}{32} \times \frac{N}{32} \times 1024$. That is, the spatial resolution is reduced by a factor of $32$, while the temporal dimension is preserved, and the number of channels is $1024$. Our network architecture is largely based on S3D-G~\cite{xie2018rethinking}, a state-of-the-art action recognition model. There are two differences between our base model and the original S3D-G model: ($i$) In our model, temporal strides for all the max pooling layers are set to 1, to leave the input's temporal dimension constant; ($ii$) We perform max spatial pooling and average temporal pooling over the resulting space-time features, as opposed to only average pooling in S3D-G.

We then collapse the temporal and spatial dimensions into a single channel. Intuitively, we want the prediction to be determined by the most dominant spatially moving object, whereas temporally, we would like to take into account the motion through the entire video segment to avoid sensitivity to instantaneous ``spiky'' motions. We therefore reduce the spatial dimension by applying global max pooling, and reduce the temporal dimension by applying global average pooling. This results in a $1024D$ vector, which is then mapped to the final logits by  a $1 \times 1$ convolution $W$. Our model is trained using a binary cross entropy loss. 

\section{Adaptive video speedup}
\label{subsection_speedup_scores}
We use our model to adaptively speed up a test video $v$. The idea is: we speed up a video. As long as the network thinks that a segment in the resulting video is \emph{not} sped-up, we can keep on speeding up that video segment even further.

\subsection{From predictions to speedup scores}
Given an input video $v$, we first generate a set of sped-up videos, $v_0, \dots v_k$, by sub-sampling $v$ by an exponential factor of $\{X^i\}_{i=0}^k$, 
where $v_0$ is the original video. We used $X=1.25$ and $k=10$ in our experiments. 
We feed each video $v_i$ into SpeedNet in a sliding window fashion; the network's prediction for each window is assigned to the middle frame. This results in a temporally-varying prediction curve, $P_{v_i}(t)$, for each video $v_i$. Here, $P_{v_i}(t)$ represents the (softmax) probability of normal speed. That is, $P_{v_i}(t) \approx 1$ if SpeedNet's prediction for a window centered at $t$ is normal speed, and $P_{v_i}(t) \approx 0$ if sped-up.

%
%

The set of speediness predictions $\{P_{v_i}\}$ are first linearly interpolated (in time) to the temporal length of the longest curve ($P_{v_0}$). We then binarize the predictions using a threshold $\rho$ to obtain a \emph{sped-up} or \emph{not sped-up} classification per timestep, denoted by the set $\{P_{v_i}^{\rho}\}$. Each vector in this set of binary speediness predictions is then multiplied by its corresponding speedup factor $X^i$ to obtain a set of speedup vectors $\{V_i(t)\}$, which are combined into a single speedup vector $V(t)$ by taking the maximum value at each timestep. In other words, $V(t)$ contains the maximum possible speedup for each timestep that was still classified as \emph{not sped-up}. The locally-adaptive speedups determine the \emph{overall} speedup of the video that still seems ``natural''. 

\subsection{Optimizing for adaptive speedup}
\label{sec:adaptive_speedup}
Our main idea is to non-uniformly change the playback speed of a video based on its content. The motivation is similar to variable bitrate (VBR) encoding, where the bandwidth allocated to a data segment is determined by its complexity. The intuition is similar---some video segments, such as those with smooth, slow motion, can be sped up more than others without corrupting its ``naturalness''.

How do we choose the threshold $\rho$, and how do we guarantee that we achieve the final \emph{desired} overall speedup with least distortions? We test for nine thresholds: $\rho \in \{0.1, \dots 0.9\}$, and select the one for which the overall speedup is closest to the desired one.

Given the per-frame speedup vector $V(t)$, as described above, our goal now is to estimate a smoothly varying speedup curve $S(t)$, which meets the user-given target speedup rate over the entire video. The motivation behind this process is that the speedup score of segments with little action will be high, meaning a human is less likely to notice the difference in the playback speed in those segments. We formulate this using the following objective: 
{\small 
\begin{equation*}
    \begin{array}{c}
         \argmin_S E_{\text{speed}}(S, V) + \beta E_{\text{rate}}(S, R_o) + \alpha E_{\text{smooth}}(S'),
    \end{array}
\end{equation*}}
where $E_{\text{speed}}$ encourages speeding frames according to our estimated speedup score $V$. $E_{\text{rate}}$ constrains the overall speedup over the entire video to match the user desired speedup rate $R_o$. $ E_{\text{smooth}}$ is a smoothness regularizer, where $S'$ denotes the first derivative of $S$. Please see Appendix \ref{speedup_opt} for full derivation and details. We then plugin the optimal speedup $S^*$ to adaptively play the video.

The graphs in Fig.~\ref{fig:adaptive_speedup} depict an example \emph{``speediness curve''} (red), along with its corresponding final optimal speedup curve $S^*$ (blue) for an \emph{overall} target speedup of $2 \times$. We define a video's \emph{``speediness curve''} to be $1-\hat{V}(t)$, where $\hat{V}(t)$ is computed by normalizing $V(t)$ to be in the range $[0, 1]$.

\section{Experiments}

\begin{figure*}
\vspace{-.2in}
\centering
    \includegraphics[width=0.95\linewidth, clip]{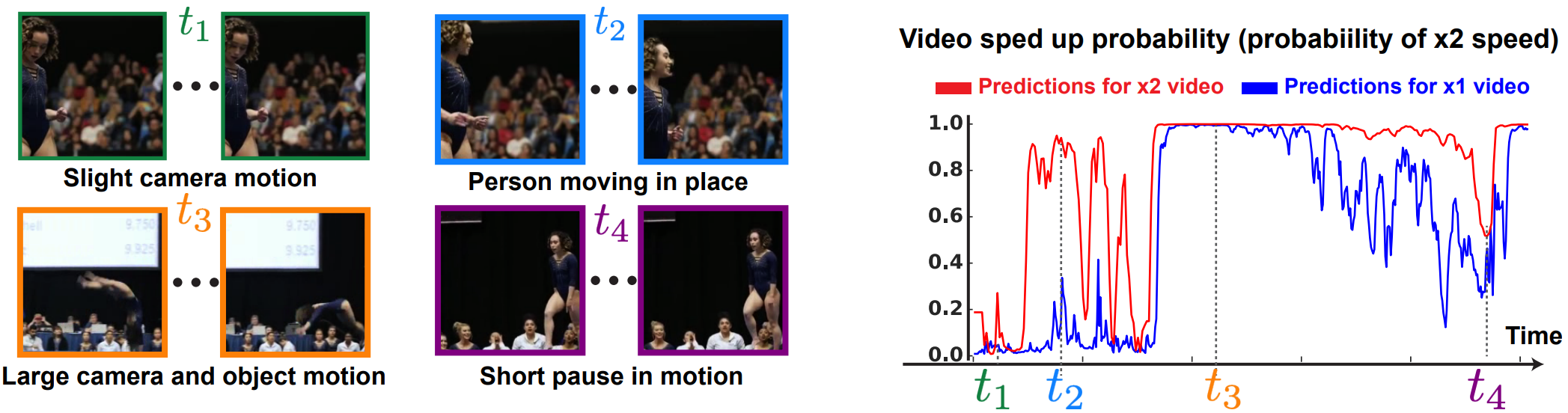} \\
    (a) ~~~~~~~~~~~~~~~~~~~~~~~~~~~~~~~~~~~~~~~~~~~~~~~~~~~~~~~~~~~~~~~~~~~~~~~~~~~~~~~~~~~~~~~~~~~~~~~~~~ (b)
\caption{
{\bf Illustration of the network's predictions.} (a) different segments from the same gymnast video are shown. (b) softmax probabilities of being ``sped-up" (y axis) are displayed for the normal speed gymnast video (blue curve) and for the sped-up gymnast video (red curve).  The segments shown in (a) are positioned in the graph. See further details in Sec.~\ref{sec:prediction_curves}.  }
  \label{fig:slowregularfast}
\afterfigure  
\end{figure*}

For our experiments we use the Kinetics~\cite{kay2017kinetics} dataset which consists of $246K$ train videos and $50K$ test clips played at 25 fps. As our method is self-supervised, we do not use the action recognition labels at training. 
We also test our model on the Need for Speed dataset (NFS)~\cite{kiani2017need} which consists of $100$ videos captured at 240 fps (a total of $380K$ frames). The dataset contains many different object actions such as moving balls, jumping people, skiing, and others. Testing our model's performance on this dataset is important for evaluating our model's generalization capability. 

\subsection{SpeedNet performance} 
\label{sub:speedup_predictions}

\begin{table}[t]
\small
\centering
\begin{tabular}{lll|cc}
\toprule[1.5pt]
\multicolumn{3}{c}{Model Type} & \multicolumn{2}{c}{Accuracy} \\
\centering
\textbf{Batch} & \textbf{Temporal} & \textbf{Spatial} & \textbf{Kinetics} & \textbf{NFS}  \\
\midrule[1pt]
Yes   & Yes      & Yes     &   75.6\%   & 73.6\%   \\
No    & Yes      & Yes     &    88.2\%  & 59.3\%    \\
No    & No       & Yes     &    90.0\%  & 57.7\%   \\
No    & No       & No      &    96.9\%  & 57.4\%    \\
\midrule[1pt]
& Mean Flow &  & 55.8\% & 55.0\%  \\
\bottomrule[1.5pt] \\
\end{tabular}
\caption{
{\bf Ablation study.} We consider the effect of spatial and temporal augmentations described in Sec.~\ref{sec:method} on the accuracy of SpeedNet for both Kinetics and NFS datasets. We also consider the effect of same-batch training (``Batch'' in the table) vs. training only with random normal-speed and sped-up video clips in the same batch (see Sec.~\ref{sec:cues}). In the last row, we consider the efficacy of training a simple network with only the mean flow magnitude of each frame.}
\label{tb:ablation}
\afterfigure
\end{table}

We assess the performance of SpeedNet (its ability to tell if a video is sped-up or not) on the test set of Kinetics and videos from the NFS dataset. We consider segments of $16$ frames either played at normal speed (unchanged) or uniformly sped-up. 
For the NFS dataset, videos are uniformly sped-up up by $10\times$ or $20\times$ to give an effective frame rate of $24$ fps (normal speed) or $12$ fps (sped-up). As the frame rate of Kinetics videos is 25 fps, we expect these speedups to correspond to $1\times$ and $2\times$ speedups of Kinetics videos. The slight variation in frame rate is important for assessing the performance of our model on videos played at slightly different frame rates than trained on. 
At test time, we resize frames to a height of $224$ keeping the aspect ratio of the original video, and then apply a $224 \times 224$ center crop. No temporal or spatial augmentation is applied. 

In Tab.~\ref{tb:ablation}, we consider the effect of training with or without: (1) temporal augmentations (2) spatial augmentations and (3) same-batch training (see Sec.~\ref{sec:cues}). When not training in ``same-batch training" mode, each batch consists of random normal-speed and sped-up video clips. 
When training SpeedNet without (1),  (2) and (3), SpeedNet relies on learned ``shortcuts''--artificial cues that are present in Kinetics, which lead to a misleading high test accuracy. However, when tested on NFS, such cues are not present, and so the test accuracy drops to 57.4\%. When using (1), (2) and
(3), reliance on artificial cues is reduced significantly (accuracy drops from 96.9\% to 75.6\%), and the gap between the test accuracy of Kinetics and that of NFS drops
to 2\%, indicating better generalization. 
While chance level is $50\%$, recall that we do not expect SpeedNet to achieve accuracy close to $100\%$, as in many cases one cannot really tell if a video is sped-up or not (e.g., when there is no motion in the clip). 

\vspace{-1em}
\subsubsection{Prediction curves}
\label{sec:prediction_curves}

Fig.~\ref{fig:slowregularfast} illustrates the predictions for a normal speed and a sped-up version of a gymnast video (more prediction results are on our \texttt{\href{http://speednet-cvpr20.github.io}{project page}}).
The predictions on the video played at normal speed ($1\times$) are shown in blue and for sped-up ($2\times$) in red. For a frame $t$, the predictions shown are $1 - P_{v_0}(t)$ ($1\times$) and $1 - P_{v_3}(t)$ ($2\times$), as detailed in Sec.~\ref{subsection_speedup_scores}. In particular, the predictions for $2\times$ are linearly interpolated so as to be displayed on the same temporal axis.
As can be seen, the region of slight camera motion ($t1$) is determined to be of normal speed for both $1\times$ and $2\times$. A person moving in place ($t2$) is determined to be sped-up for $2\times$ and of normal speed for $1\times$. Large camera and object motion ($t3$) is determined to be sped-up for both $1\times$ and $2\times$. Lastly, a short pause in motion ($t4$), has roughly equal probability of being sped-up and of being normal speed for both $1\times$ and $2\times$. 

\vspace{-1em}
\subsubsection{Comparison to optical flow}

We consider the efficacy of training a baseline model whose input is the per-frame average flow magnitude for each example in our Kinetics training set. This results in a vector of size $T$ for each video clip. We train a simple network with two fully connected layers, $ReLU$ activations and batch normalization. As can be seen in Tab.~\ref{tb:ablation}, this model achieves only $55\%$ accuracy on the test sets of Kinetics and NFS. A major limitation of the mean optical flow is its correlation with the distance of the object from the camera, which can be seen in Fig. \ref{fig:speediness_neq_motion_magnitude}. While SpeedNet is clearly superior compared to the flow baseline, it does tend to fail in scenarios which contain either extreme camera motion or very large object motion, such as fast moving objects and motion very close to the camera. We hypothesize that this is because our training set does not contain enough normal-speed videos with such large frame-to-frame displacements, which is usually characteristic of videos being played at twice their original speed.


\subsection{Generalization to arbitrary speediness rates}
We tested our model on a variety of real-world videos downloaded from the web which contain slow motion effects, involving natural camera motion as well as complex human actions, including ballet dancing, Olympic gymnastics, skiing, and many more. Our algorithm manages to accurately predict which segments are at normal speed and which are \emph{slowed down} by using the method described in Sec.~\ref{subsection_speedup_scores}. To emphasize, even though our SpeedNet model was trained on a dataset of videos at normal speed and $2\times$ speed, we can use it within our framework to classify video segments which contain slow-motion, i.e. whose duration at playback is slower than real-time. A video clip is determined as ``slow-motion'' if its sped-up version is detected as ``normal-speed'', such as the example shown in Fig.~\ref{fig:teaser}.



\subsection{Adaptive speedup of real-world videos}
\label{sub:adaptive_results}

To evaluate our adaptive speedup against that of uniform speedup, we seek videos where there is a large difference in the ``speediness" of objects within the video. For example, for a $100m$ sprint run, the runner is initially walking towards the running blocks, then not moving at all just before the sprint (when at the blocks), and finally sprinting. 
We performed adaptive speedup on five such videos from YouTube, and then conducted a user study to determine the objective quality of our results. For each video, our adaptive speedup and its corresponding uniformly sped-up version are shown to the user at random, who is asked to select the sped-up version which ``looks better''. 

\begin{figure*}
\centering
    \includegraphics[width=0.97\linewidth, clip]{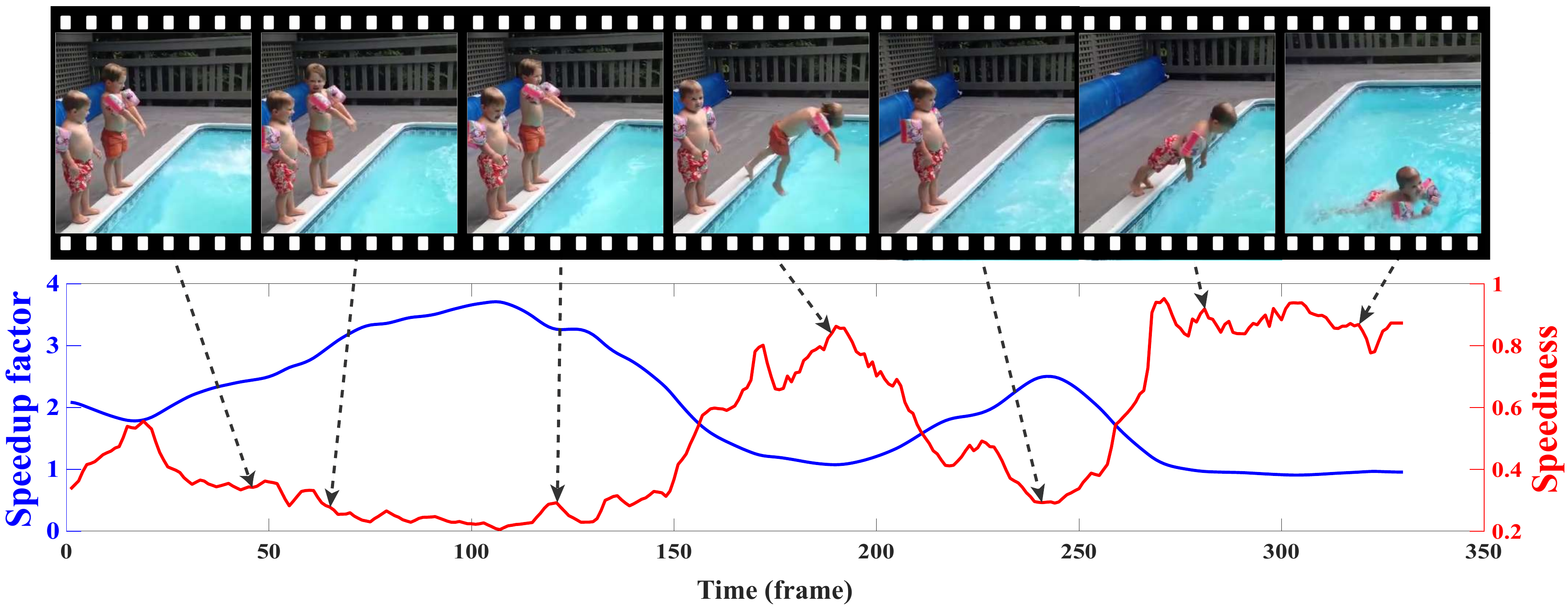}
\caption{{\bf Adaptive video speedup.} We apply the SpeedNet model for generating time-varying, adaptive speedup videos, based on the frames' speediness curve (Sec.~\ref{subsection_speedup_scores}). Here we show the speediness curve and our resulting adaptive speedup factor for a video of two kids jumping into a pool. Several selected frames are shown at the top, pointing to their corresponding times within the sequence on the predicted speediness curve.}
 \label{fig:adaptive_speedup}
\afterfigure
\end{figure*}

\begin{figure}[t]
\centering
    \includegraphics[width=\columnwidth]{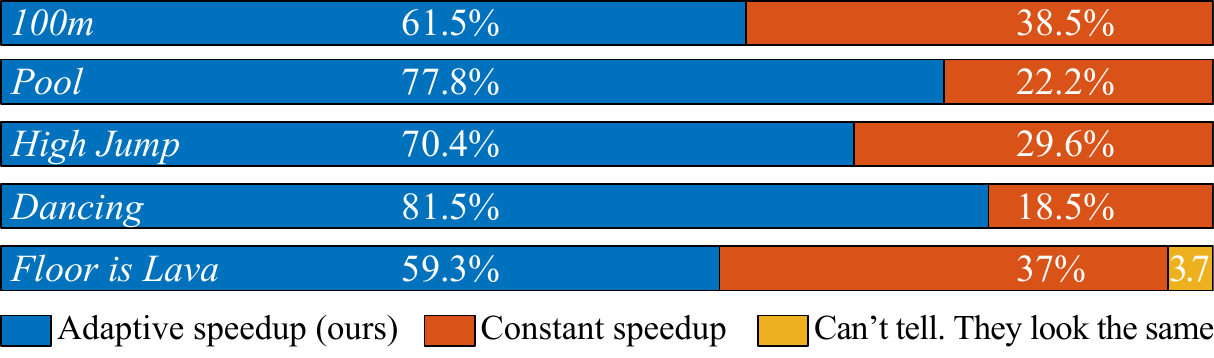}
\caption{{\bf Adaptive video speedup user study.} We asked $30$ participants to compare our adaptive speedup results with constant uniform speedup for $5$ videos (without saying which is which), and select the one they liked better. Our adaptive speedup results were consistently (and clearly) preferred over uniform speedup.}
 \label{fig:adaptive_speedup_user_study}
\afterfigure
\end{figure}

We conducted the study on $30$ users with different research backgrounds, and for all five videos we presented, our adaptive speedup was preferred by a clear margin over uniform speedup, as shown in Fig.~\ref{fig:adaptive_speedup_user_study}. 
An example of one of the adaptively sped-up videos used in our study is shown in Fig.~\ref{fig:adaptive_speedup}, and all five videos are on our \texttt{\href{http://speednet-cvpr20.github.io}{project~page}}.

\subsection{SpeedNet for self-supervised tasks}
Solving the speediness task requires high-level reasoning about the natural movements of objects, as well as understanding of lower-level motion cues. Since SpeedNet is self-supervised,
we evaluate the effectiveness of its internal representation on the self-supervised tasks of pre-training for action recognition and video retrieval. 

\vspace{-1em}
\subsubsection{Action recognition}
Utilizing self-supervised pre-training to initialize action recognition models is an established and effective way for evaluating the internal representation learned via the self-supervised task. A good initialization is important especially when training on small action recognition datasets, such as UCF101~\cite{soomro2012ucf101} and HMDB51~\cite{hmdb51}, as the generalization capability of powerful networks can easily be inhibited by quickly overfitting to the training set.

Fine-tuning a pre-trained SpeedNet model on either UCF101 or HMDB51 significantly boosts the action recognition accuracy over random initialization, which indicates that solving the speediness task led to a useful general internal representation. In Tab.~\ref{tb:actrec} we show that our action recognition accuracy beats all other models pre-trained in a self-supervised manner on Kinetics. For reference, we include the performance of S3D-G network when pre-trained with ImageNet labels (ImageNet inflated), and when using the full supervision of Kinetics (Kinetics supervised). Both networks use additional supervision, which we do not.

The best performing self-supervised model we are aware of, DynamoNet \cite{diba2019dynamonet}, was pre-trained on YouTube-8M dataset \cite{yt8m}, which is $10\times$ larger than Kinetics, and whose raw videos are not readily available publicly. DynamoNet attains accuracies of 88.1\% and 59.9\% on UCF101 and HMDB51, respectively. 

Note that our strong random init baselines for S3D-G are in part due to using $64$ frames during training. SpeedNet was designed and trained for the specific requirements of ``speediness'' prediction and, as such, was not optimized for action recognition. 
For reference, when trained with a weaker architecture such as I3D~\cite{i3d}, our speediness prediction drops to $63.1\%$, but we observe a larger absolute and relative improvement over the random init baseline for both datasets, as reported in Tab.~\ref{tb:actrec}. 

\begin{table}
\small
\centering
\begin{tabular}{llcc}
\toprule[1.5pt]
\multicolumn{2}{c}{Initialization} & \multicolumn{2}{c}{Supervised accuracy} \\
\bf Method & \bf Architecture & \bf UCF101 & \bf HMDB51 \\
\midrule[1pt]
Random init & S3D-G & 73.8 & 46.4 \\
ImageNet inflated & S3D-G & 86.6 & 57.7 \\
Kinetics supervised & S3D-G & 96.8 & 74.5 \\
\midrule
CubicPuzzle~\cite{kim2019self} & 3D-ResNet18 & 65.8 & 33.7 \\
Order~\cite{xu2019self} & R(2+1)D & 72.4 & 30.9 \\
DPC~\cite{han2019video} & 3D-ResNet34 & 75.7 & 35.7 \\
AoT~\cite{wei2018learning} & T-CAM & 79.4 & -   \\
SpeedNet (Ours) & S3D-G & \bf 81.1 & \bf 48.8 \\
\midrule
Random init & I3D & 47.9 & 29.6 \\
SpeedNet (Ours) & I3D & 66.7 & 43.7 \\
\bottomrule[1.5pt] \\
\end{tabular}
\caption{{\bf Self-supervised action recognition.} Comparison of self-supervised methods on UCF101 and HMDB51 split-1. The top methods are baseline S3D-G models trained using various forms of initialization. All of the methods in the middle were trained with a self-supervised method on Kinetics and then fine-tuned on UCF101 and HMDB51. On the bottom, we show for reference our random init and SpeedNet accuracy when trained on I3D network. }
\label{tb:actrec}
\afterfigure
\end{table}

\vspace{-1em}
\subsubsection{Nearest neighbor retrieval}
Another way of assessing the power of SpeedNet's learned representation is by extracting video clip embeddings from the model, and searching for nearest neighbors in embeddings space. In particular, given a video clip of arbitrary spatial dimension and temporal length, we propose to use the max and average-pooled space-time activations, described in Sec.~\ref{sec:arch}, as a $1024D$ feature vector representing the clip. The experiments described in this section demonstrate that the extracted features encapsulate motion patterns in a way that facilitates the retrieval of other clips with similar behavior.

\begin{figure}[t]
\vspace{-.1in}
    \centering
    \includegraphics[width=.95\columnwidth]{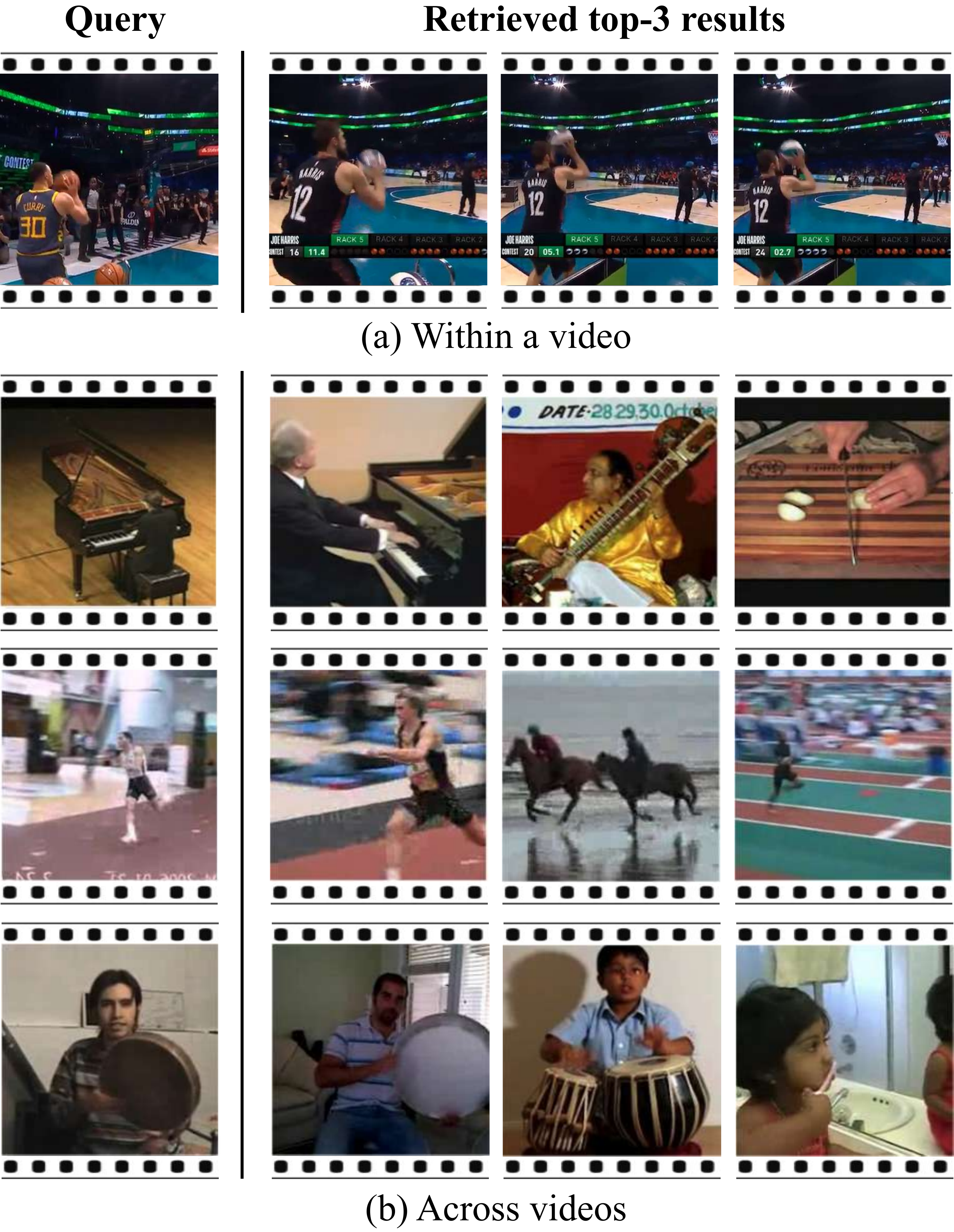}
    \caption{{\bf Video clip retrieval.}
    The left column shows an image from a query clip, and the right three columns show the clips with the closest embeddings. In (a), the retrieval is from clips taken from further along in the same video. In (b), the results are retrieved from the entire UCF101 train set. Note that the embeddings focus more on the type of movement than the action class, for example the girl in the last row is making similar back/forth/up/down motions with her hand as the hand drummer in the query.}
    \label{fig:nn}
\vspace{-1.5em}    
\end{figure}

In this experiment, we extract a 16-frame sequence from a video (\emph{query}, see Fig.~\ref{fig:nn}), and aim to retrieve similar clips from a either the same (longer) video (``within a video''), or from a collection of short video clips (``across videos''). For the former, we first extract the query feature vector from SpeedNet, and proceed to calculate feature vectors from the target video in sliding-window fashion over 16-frame windows. We then calculate a similarity graph by computing the cosine similarity score between the query feature and each of the target video features. In the first experiment, the query is of a basketball player shooting a 3-point shot, and similar clips of a different player are retrieved from further along in the same video, filmed from a slightly different angle and scale. In Fig.~\ref{fig:nn} (a) we show a representative frame from each peak in the similarity graph.

In the second experiment (Fig.~\ref{fig:nn} (b)), the query clip is taken from the test set of UCF101, and we search for the nearest neighbors in the training set, again using cosine similarity. SpeedNet mostly focuses on the type and speed of an object's behavior, which is not always equivalent to the class of the video. For example, in Fig.~\ref{fig:nn}, the girl in the last row is making similar back/forth/up/down motions with her hand as the hand drummer in the query. 

We would, however, like to measure how correlated our learned representations are to specific action classes. We consider a third experiment where we measure the Recall-at-topK: a test clip is considered correctly classified if it is equivalent to the class of one of the $K$ nearest training clips. We use the protocol of Xu \emph{et al.}~\cite{xu2019self} (denoted \emph{Order}). As can be seen in Tab.~\ref{tb:recallatk}, our method is competitive with other self-supervised methods and only slightly inferior to \cite{xu2019self}.

\begin{table}
\small
\centering
\begin{tabular}{p{1.2cm}p{1.4cm}ccccc}
\toprule[1.5pt]
\bf Method & \bf Architecture & \bf 1 & \bf 5 & \bf 10  & \bf 20 & \bf 50\\
\midrule[1pt]
Jigsaw~\cite{jigsaw} & CFN  &  19.7 & 28.5 & 33.5 & 40.0 & 49.4 \\
OPN~\cite{opn} & OPN  & 19.9 & 28.7 &  34.0 &  40.6 &  51.6 \\
Buchler~\cite{bucher} & CaffeNet & 25.7 &  36.2  & 42.2 &  49.2 &  59.5 \\
Order~\cite{xu2019self} & C3D & 12.5 & 29.0 & 39.0 & 50.6 &  66.9 \\
Order~\cite{xu2019self} & R(2+1)D & 10.7 & 25.9 &  35.4 &  47.3 &  63.9 \\
Order~\cite{xu2019self} & R3D & 14.1 & 30.3 & 40.0 & 51.1 & 66.5 \\
\midrule[1pt]
Ours & S3D-G & 13.0 & 28.1 & 37.5 &  49.5 & 65.0 \\
\bottomrule[1.5pt] \\
\end{tabular}
\caption{{\bf Recall-at-topK.} Top-K accuracy for different values of K for UCF101.}
\label{tb:recallatk}
\afterfigure
\end{table}

\subsection{Visualizing salient space-time regions}
\label{sec:visualization}

To gain a better understanding about which space-time regions contribute to our predictions, we follow the Class-Activation Map (CAM) technique \cite{zhou2016learning} to visualize the energy of the last 3D layer, before the global max and average pooling (see Fig.~\ref{fig:speednet}). More specifically, we extract a $T \times N \times N \times 1024$ feature map, where $T$ and $N$ are the temporal and spatial dimensions, respectively. We first reduce the number of channels to $T \times N \times N $ using $W$ (the weights that map from the $1024D$ vector to the final logits). We then take the absolute value of the activation maps and normalize them between $0$ and $1$. 

\begin{figure}[t]
\centering
    \includegraphics[width=\columnwidth]{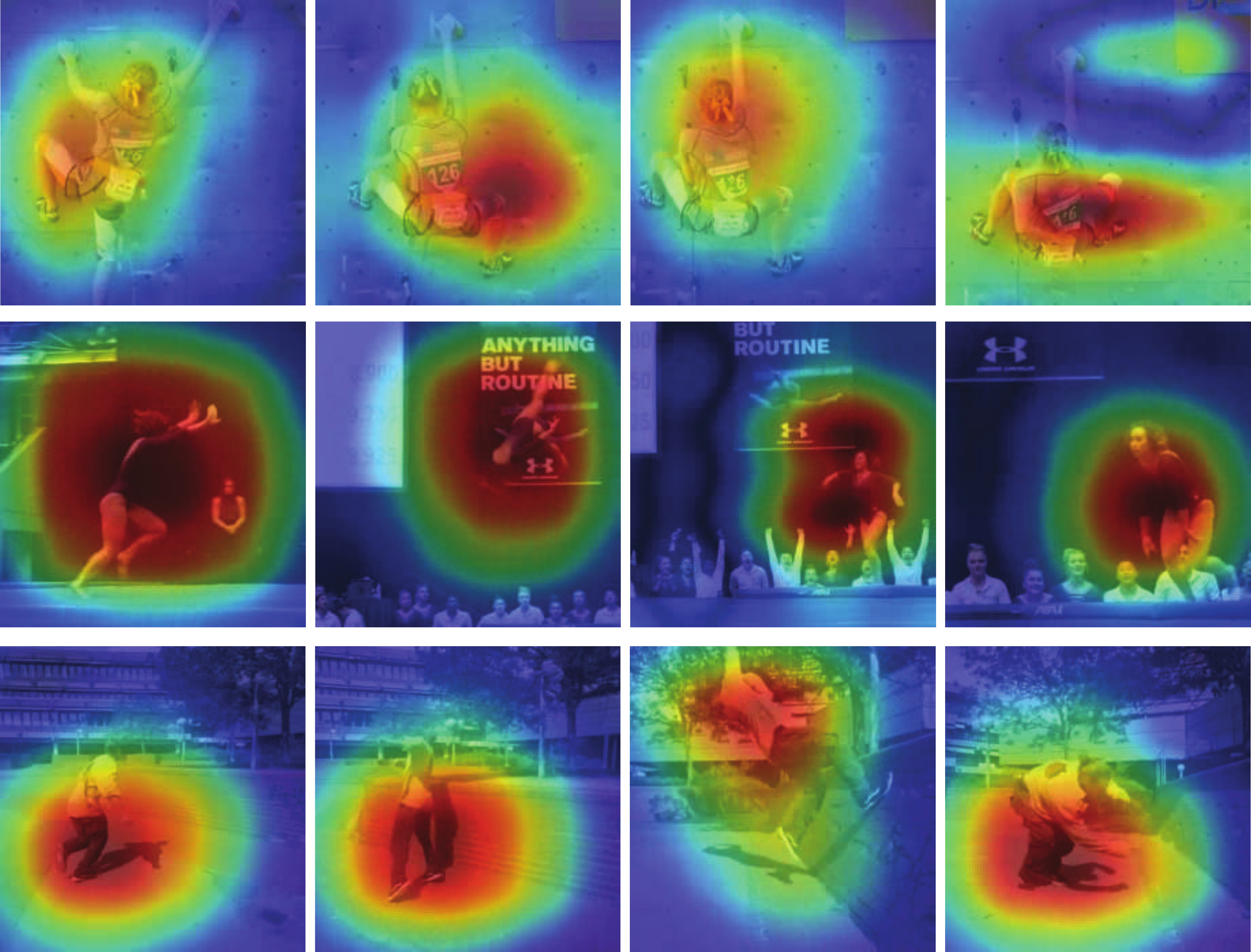}
\caption{{\bf Which space-time regions contribute the most to our speediness predictions?} CAM visualizations as detailed in Sec.~\ref{sec:visualization}. We visualize such regions as overlaid heat-maps, where red and blue correspond to high and low activated regions, respectively. Interestingly, the model is able to pick up the salient mover in the presence of significant camera motion.}
  \label{fig:heatmaps}
\afterfigure  
\vspace{0.5em}
\end{figure}

Fig.~\ref{fig:heatmaps} depicts computed heat maps superimposed over sample frames from several videos are shown in . These examples portray a strong correlation between highly activated regions and the dominant mover in the scene, even when performing complex actions such as flips and articulated motions. 
For example, in the top row, second frame, the network attends to the motion of the leg, and in the second row, activations are highly concentrated on the body motion of the gymnast. 
Interestingly, the model is able to pick up the salient mover in the presence of significant camera motion. 

\begin{figure}[t]
    \centering
    \includegraphics[width=\columnwidth]{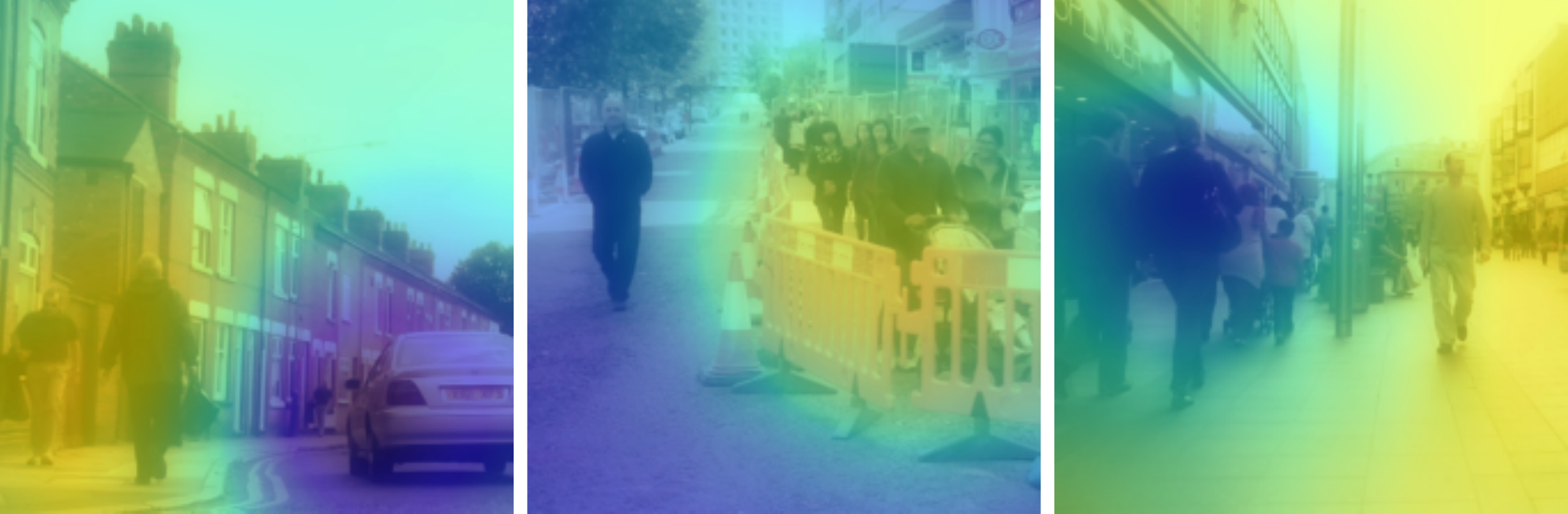}
    \caption{{\bf Spatially-varying speediness.} In the video ``Memory Eleven''~\cite{memory_eleven}, one part of the frame is played in slow motion while the other part is played at regular speed. Using CAM visualizations without taking the absolute value (thus maintaining the direction of the activations, see Sec.~\ref{sec:visualization}), we can see that the model accurately predicts the speediness of each part of the frame part, from blue (normal speed), to red (slow motion).}
    \label{fig:heatmaps_different_regions}
\afterfigure    
\end{figure}

In Fig.~\ref{fig:heatmaps_different_regions}, we consider the video ``Memory Eleven"~\cite{memory_eleven}, in which part of the frame is played in slow motion while the other part is played at normal speed. We use a similar visualization as in Fig.~\ref{fig:heatmaps}, but do not take the absolute value of the activation maps. While in Fig.~\ref{fig:heatmaps} we are interested in overall important spatio-temporal regions for classification, in Fig.~\ref{fig:heatmaps_different_regions} we are interested in distinguishing between areas in the frame used for classifying the video as normal and as sped-up. SpeedNet accurately predicts the speediness of each part, from blue (normal speed), to red (slow motion). 


\section{Conclusion}

Our work studies the extent to which a machine can learn the ``speediness'' of moving objects in videos: whether an object is moving slower, at, or faster than its natural speed. To this end, we proposed SpeedNet, a model trained in a self-supervised manner to determine if a given video is being played at normal or twice its original speed. We showed that our model learns high level object motion priors which are more sophisticated than motion magnitude, and demonstrated the effectiveness of our model for several tasks: adaptively speeding up a video more ``naturally'' than uniform speedup; as self-supervised pre-training for action recognition; as a feature extractor for video clip retrieval.




{\small
\bibliographystyle{ieee_fullname}
\bibliography{references}
}

\vfill\null

\appendix
\section{Adaptive speedup optimization}
\label{speedup_opt}


The term  $E_{\text{speed}}$ defines to which extent each frame is sped up according to our estimated speedup score $V$:
\begin{small}
\begin{align}
\label{eq:optim}
         E_{\text{speed}}(S, V) &= \sum_t (1-\hat{V}(t)) (S(t) - R_{min})^2 \\ & + \gamma \sum_t \hat{V}(t)(S(t) - R_o)^2 \nonumber
\end{align}
\end{small}
The first term of Eq.~\ref{eq:optim} encourages the estimated speedup to be close to a low speedup rate ($R_{min}$) for frames whose speediness score is high ($1-\hat{V}(t)$ is close to one). The second term encourages the estimated speedup to be close to the total desired rate $R_o$, and is higher for frames whose speediness score is low (hence can be sped up more); this term mostly serves as a regularizer, and prevents high spikes in the estimated speedup curve. 

$E_{\text{rate}}$ constrains the overall speedup over the entire video to match the user desired speedup rate $R_o$, and is given by:
{\small 
\begin{align}
         E_{\text{rate}}(S, R_o)= \left(\frac{1}{T} \sum S(t) - R_o \right)^2
         \end{align}}
$E_{\text{smooth}}$ is a smoothness regularizer given by:
{\small 
\begin{align}
         E_{\text{smooth}}(S')= \sum_t \left(S(t) - S(t+1)\right)^2
         \end{align}}
This optimization problem has a closed-form solution, where the optimal speedup curve $S^*$ is given by a solution to a linear system. In practice, for efficiency, we used Tensorflow optimizer to minimize it. The playback of the video is then adaptively changed according to the optimal speedup curve $S^*$ (see Fig.~\ref{fig:adaptive_speedup} (blue)).

\end{document}